\documentclass[letterpaper, 10 pt, conference]{ieeeconf}  % Comment this line out if you need a4paper

\IEEEoverridecommandlockouts                              % This command is only needed if 
\usepackage{graphics} % for pdf, bitmapped graphics files
\usepackage{epsfig} % for postscript graphics files
\usepackage{times} % assumes new font selection scheme installed
\usepackage{amsmath} % assumes amsmath package installed
\usepackage{amssymb}  % assumes amsmath package installed
\usepackage{algorithmic}
\usepackage{algorithm}
\usepackage[backend=bibtex]{biblatex}
\usepackage{pseudo}
\usepackage{xcolor}
\usepackage{amsmath}
\usepackage{hyperref}
\hypersetup{
     colorlinks   = true,
     citecolor    = green
}

\title{\LARGE \bf
Shaped Policy Search for Evolutionary Strategies using Waypoints*
}

\author{Kiran Lekkala$^{1}$ and Laurent Itti$^{2}$% <-this % stops a space
\thanks{*This work was supported by C-BRIC (one of six centers in JUMP, a Semiconductor Research Corporation (SRC) program sponsored by DARPA), the Army Research Office (W911NF2020053), and the Intel and CISCO Corporations. The authors affirm that the views expressed herein are solely their own, and do not represent the views of the United States government or any agency thereof.}% <-this % stops a space
\thanks{$^{1}$Kiran Lekkala is with the ILab, Department of Computer Science,
        University of Southern California, 90089, USA
        {\tt\small klekkala@usc.edu}}%
\thanks{$^{2}$Laurent Itti is with the ILab, Department of Computer Science, Psychology and NGP, University of Southern California
        {\tt\small itti@usc.edu}}%
}

\bibliography{main}

\begin{document}

\maketitle
\thispagestyle{empty}
\pagestyle{empty}

%%%%%%%%%%%%%%%%%%%%%%%%%%%%%%%%%%%%%%%%%%%%%%%%%%%%%%%%%%%%%%%%%%%%%%%%%%%%%%%%
% computational cost, time-complexity, generation vs iteration, how much time is used in train vs testing
% comprehensive comparisons with non-evolutionary based approaches
% what if intermediate sub-goals are not available
% Discussion and Conclusion section is too brief and more discussions are expected
%
%%%%%%%%%%%%%%%%%%%%%%%%%%%%%%%%%%%%%%%%%%%%%%%%%%%%%%%%%%%%%%%%%%%%%%%%%%%%%%%%

\begin{abstract}

In this paper, we try to improve exploration in Blackbox methods, particularly Evolution strategies (ES), when applied to Reinforcement Learning (RL) problems where intermediate waypoints/subgoals are available. Since Evolutionary strategies are highly parallelizable, instead of extracting just a scalar cumulative reward, we use the state-action pairs from the trajectories obtained during rollouts/evaluations, to learn the dynamics of the agent. The learnt dynamics are then used in the optimization procedure to speed-up training. Lastly, we show how our proposed approach is universally applicable by presenting results from experiments conducted on Carla driving and UR5 robotic arm simulators.

\end{abstract}

\begin{figure}
    \centering
    \includegraphics[width=\columnwidth]{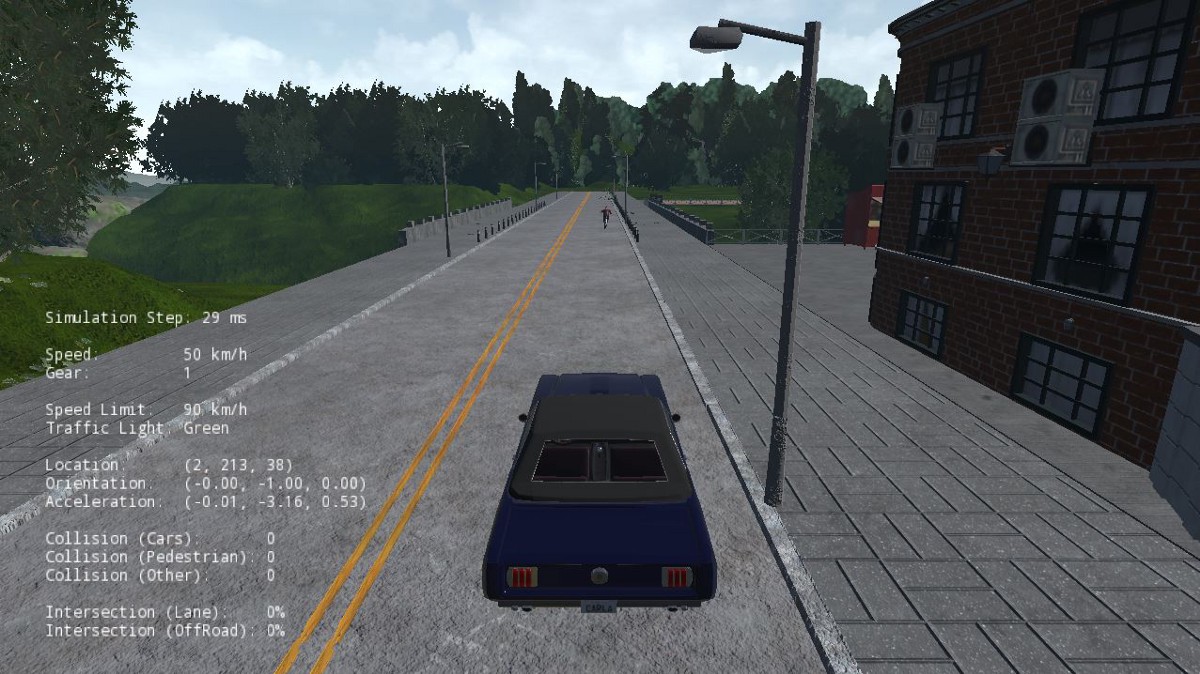}
    \includegraphics[width=\columnwidth]{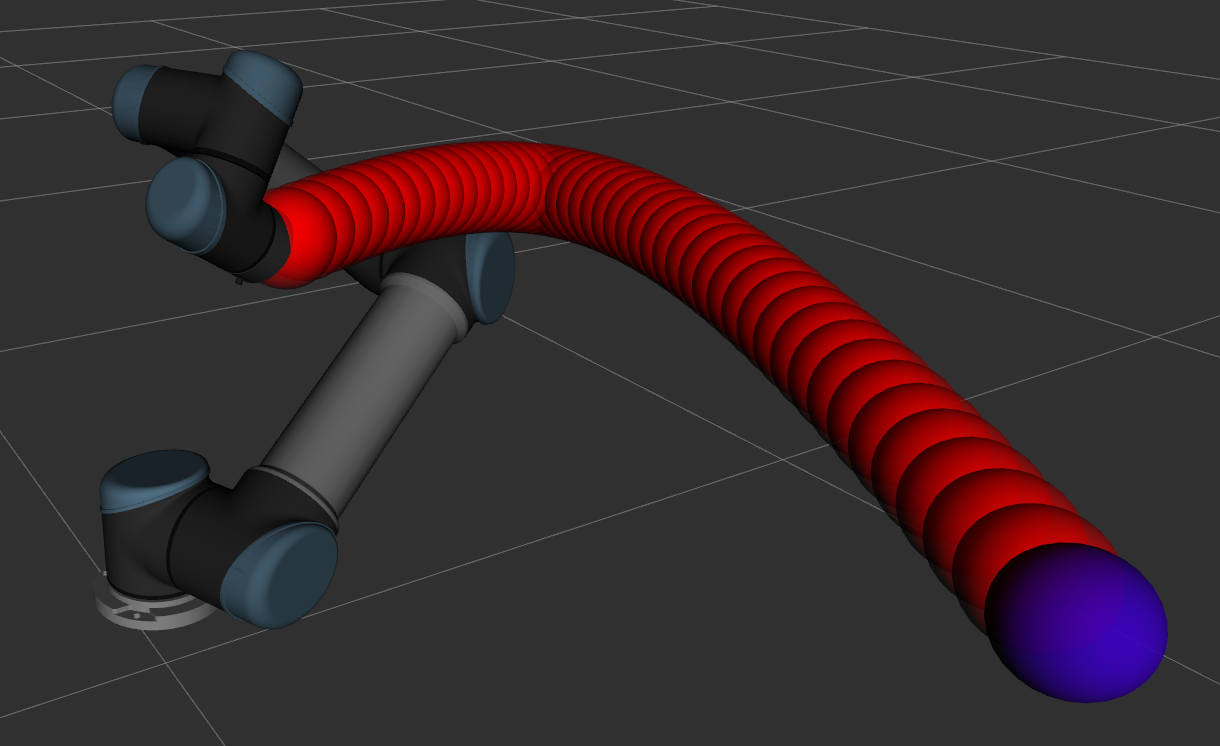}
    \caption{We test our method on the (a) Carla and the (b) UR5 Gazebo simulators. In Carla simulator, the agent had to solve a point goal navigation task, where it has to reach a destination location from a randomly spawned location. In UR5 Gazebo simulator, the task for the agent was to move the fixed end-effector in alignment with the trajectory consisting of the target pose (blue) and the waypoints (red).}
    \label{fig:intro}
\end{figure}
\section{Introduction}

%%%introduction
%what is the problem
\textit{Reinforcement learning (RL)} is one of the most popular methods used in a wide range of Robotic applications. Although RL has shown promising results in the past, one of the main challenges that arise in RL is that it requires evaluating a large number of samples on the environment. Evolutionary strategies (ES) are one of the least widely used RL methods, perhaps because, although they are easily parallelizable, they have poor sample efficiency. In situations where we have access to waypoints/intermediate subgoals, we could further improve the performance of these methods.

%why are we concerned with the problem.. plethora of prior works. .
A common practice to run RL algorithms involves single-threaded sequential policy learning. Algorithms on scaling these sequential learning methods to distributed settings \cite{DBLP:conf/iclr/WijmansKMLEPSB20, DBLP:journals/corr/ZhaoC13} are getting popular as they can be linearly scaled to the number of machines/cores. However, these distributed methods require the entire gradient of the policy to be communicated across machines, which imposes restrictions on the network bandwidth and limits usage. On the other hand, ES methods just transmit parameters and the reward value and the distributed computation is only used for rollouts.

%existing solutions to the problem.. mention broadly mention ben's work and hindsight experience replay why these methods which form a basis of 3 different aspects of your work fail in your case.
%\textit{AllReduce} \cite{DBLP:journals/corr/ZhaoC13} is the most common algorithm used to extend policy gradient methods to distributed settings. Existing solutions like \cite{DBLP:conf/rss/RajeswaranKGVST18}, \cite{DBLP:conf/nips/EysenbachSL19} and \cite{DBLP:conf/nips/AndrychowiczCRS17} better RL by including expert demonstrations, using sub-goals, and learning from failed distributions.

\begin{figure*}
    \centering
    \includegraphics[width=\textwidth]{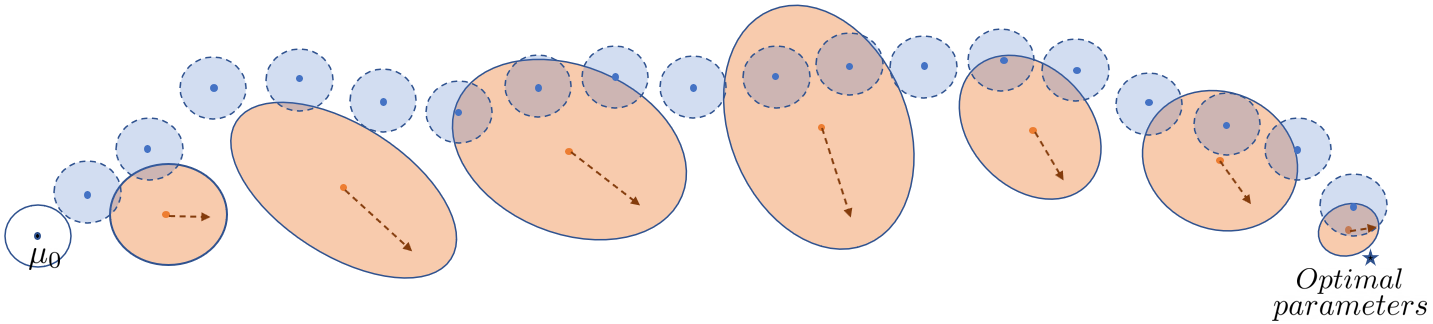}
    \caption{Pictorial overview of our method. Blue represents the uniform sampling distributions used by vanilla-ES method, and orange represents the sampling done by our method. Center dots in both the sampling distributions represents the mean. Instead of sampling parameters from a standard Gaussian of a fixed covariance, like that of vanilla-ES, we sample parameters from a Gaussian distribution obtained by extending the search space towards the optimum using the surrogate gradients (brown).}
    \label{fig:second}
\end{figure*}

%what is your solutions to the problem.. make it brief
Existing RL works focus on policy optimization and neglect learning useful information like agent dynamics, i.e., which action would lead to a specific state transition. Our method shapes the policy search such that the sampling procedure is biased towards the optimal parameters (Figure \ref{fig:second}). To achieve this, we generate a noisy action label and evaluate a surrogate/noisy estimate of the gradients of the policy parameters. We can then sample parameters based on their alignment with these gradients, unlike uniform sampling.

\section{Related Work}
%Reinforcement Learning.. rock paper scissors..direct policy search
%interesting things abou rl which are relevent to your work
Standard RL methods rely on reward signals from the environment to find the optimal parameters for a policy. Two of the most active research areas in RL are overcoming sparse rewards and exploration. In many practical scenarios, environments have long horizon and sparse rewards, and it becomes challenging for RL methods to optimize the reward function. \cite{DBLP:conf/nips/AndrychowiczCRS17} proposes a novel approach to alleviate sparse rewards by learning from failed trajectories, which dovetails with our motivation. Although shaped rewards can be used these scenarios, they often lead to sub-optimal solutions.

Derivative-free optimization methods have been used in the past for problems where computation of a gradient is not feasible, like that of RL. Most popular \textit{BlackBox methods} like evolutionary strategies work by sampling parameters from a distribution and evaluating them to take an update step. \textit{Covariance-based Evolution strategy (CMA-ES)} \cite{DBLP:journals/corr/Hansen16a} and \textit{Cross-Entropy} \cite{book:10.5555/1014902} are two most prominent algorithms widely used in RL and in they perform surprisingly well when compared to the state of the art policy gradient methods. Other versions of ES like PEPG \cite{DBLP:journals/nn/SehnkeORGPS10} and NES \cite{DBLP:journals/jmlr/WierstraSGSPS14} carry out parameter search with novel techniques inspired from Reinforce, which is one of the earliest policy gradient algorithm \cite{ha2017visual}.

%Imitation Learning.. umbrella.. significant challenge
\textit{Imitation learning (IL)} involves learning a policy from a set of expert demonstrations without explicitly designing any reward function. The most naive form of Imitation learning, sometimes referred to as \textit{Behaviour Cloning} requires fitting a model to the expert trajectories \cite{wen2022can, xu2022ferroelectric, DBLP:journals/corr/abs-2305-15591, lekkala2020attentive}. Although behaviour cloning is a seemingly trivial problem, a significant challenge involves distribution mismatch of the expert and the agent, which leads to compounding errors. Furthermore, the trained agent would have obtained sub-optimal behaviour compared to the expert, and so the agent can never be better than the expert. Recent advances like enabling the use of \textit{Inverse Reinforcement Learning (IRL)} also fall under the broad umbrella of IL. These methods use expert demonstrations to learn the reward function \cite{lekkala2019meta, lekkala2016accurate, lekkala2015artificial, lekkala2014pid, lekkala2016simultaneous}, which could then be used to optimize the policy using RL.

Expert demonstrations have also been used to supplement exploration in RL. Some of the prominent ones include fusing demonstrations in the optimization procedure with Q-learning as an auxiliary objective \cite{DBLP:journals/corr/abs-1802-05313} and hybrid policy gradient \cite{DBLP:conf/rss/RajeswaranKGVST18}. Although the methods mentioned above show promising results on some benchmarks, there are inherent issues. First, expert demonstrations are off-policy, which makes the optimization challenging. Second, while trying to learn a policy using RL, the agent could forget the expert's trajectories, which nullifies the reason for bringing in the demonstrations.

%Ifo foster reinforcement learning.. covariate shift
\textit{Imitation from Observation (IfO)} is an ongoing research area which deals with policy learning from expert observations, without the need of the expert actions. The policy is then learnt with the help of RL, by taking help from the expert observations, which could be same as the waypoints/intermediate goals. Unlike using demonstrations in RL, since there are no actions for the expert, there would not rise a problem of fusing in off-policy data. However, to determine the actions and reachable states is still a challenge. Recent IfO works, like \cite{DBLP:conf/aaai/GuoCYTC19} and \cite{DBLP:journals/ral/PavseTHWS20} use an \textit{Inverse Dynamics} model.

%Motivation Our Contribution and how is our contribution different from others.. mention the number of parameters in the experiment section are known to be highly sample-inefficient..
 
% Since these intermediate waypoints do not have any associated actions, unlike demonstrations, we generate an action label by learning dynamics of the agent and use it in improve exploration.

% Since the state vector from the simulator is composed of odometry or joint angles information 
Although Blackbox methods, like CMA, are not constrained to environments having \textit{Markov Decision Process (MDP)} assumption, and so can be used in a lot more applications, they are highly sample-inefficient. However, if we have access to intermediate waypoints, these BlackBox methods could be improved by learning an inverse dynamic model from the experiences generated. Since plenty of rollouts are generated by parallelizing across multiple machines, the inverse dynamics model could be trained efficiently.

In the ideal case, if these actions generated were the actual goal/task-specific action labels, this entire formulation would be reduced to a trivial problem of behaviour cloning. However, to obtain the action vectors is itself the problem we are trying to solve, i.e., policy estimation, we could only get the inverse dynamics model, that is conditioned on the intermediate waypoint, to infer a rough estimate of the action. We would then get noisy gradients evaluated on the data consisting of the noisy action labels, which could be used to bias the sampling process in the evolution strategy. This approach would favour the regions of the search space closer to the optimal parameters.

\begin{figure}
    \centering
    \includegraphics[width=\columnwidth]{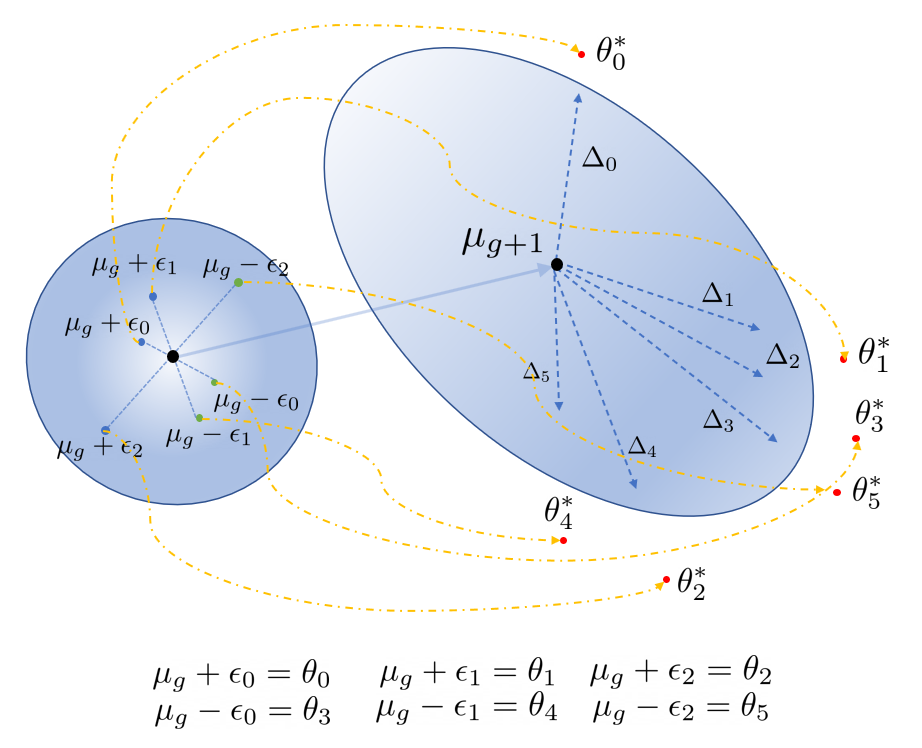}
    \caption{Above figure shows how our method changes the uniform sampling to favour samples closer to the global optimum. For simplicity, we consider population size to be 6, where half of them have positive perturbations (in blue), and the other half have negative perturbations (green). These perturbations are with respect to the mean $\mu$ (in black). After evaluating these samples on the environment, they are updated to $\theta_i^*$ (in red) using the noisy actions from the inverse-dynamics model by behaviour cloning. Yellow dot-broken arrow represents gradient descent when done by each worker on the sample. The sampling distribution is then modified using the surrogate gradients $\Delta_i$ pointing towards $\theta^*_i$.}
    \label{fig:specific}
\end{figure}

%mention inverse dynamics before this
Unlike \cite{DBLP:journals/ral/PavseTHWS20}, where the policy is the same as the inverse dynamics model, our method is more practical as the policy does not depend on any intermediate waypoints. The inverse dynamics model is only used to speed-up training and does not play any role during test-time. Following are our contributions:

We propose a methodology to extend and apply Guided Evolutionary strategy \cite{DBLP:conf/icml/Maheswaranathan19} to RL settings where the agent has access to the waypoints in an environment. This also involves, using an inverse dynamics model to generate noisy action labels, from which gradients are evaluated. In the experimental section, we outline how our method can be scaled and deployed on a Beowulf cluster.
%===============================================================================

\begin{figure*}
    \centering
    \includegraphics[width=6.1in]{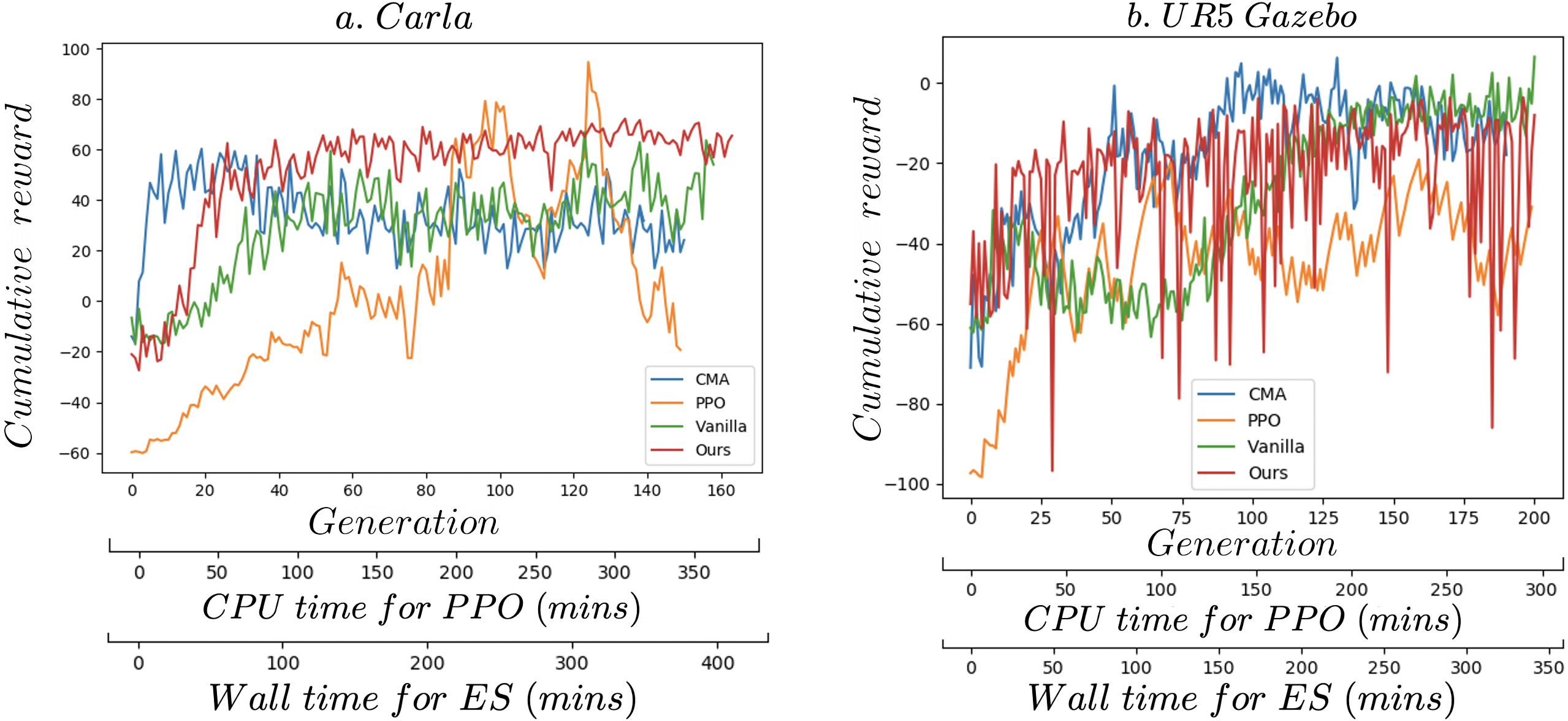}
    \caption{Comparison of our method with the benchmarks CMA-ES, Vanilla-ES, PPO on the (a) CARLA driving simulator and (b) UR5 Gazebo simulator. $x$ axis represents the generation, CPU time and the Wall time. CPU time is used for PPO which runs on a single core and wall time represents the actual time taken by the evolutionary strategy irrespective of the number of cores. $y$ axis represents the cumulative reward. We provide the mean reward of the best member in the population averaged on 15 test episodes, in the case of ES benchmarks. Our method reaches the global optimum faster and performs better than all the baselines.}
    \label{fig:benchmark}
\end{figure*}

\section{Problem formulation and preliminaries}
%rl, mdp, ppo
Reinforcement learning methods are based on an MDP formulation, characterized by the tuple $(\mathcal{S}, \mathcal{A}, \mathcal{R}, \mathcal{S^{'}})$, where $\mathcal{S}$ and $\mathcal{A}$ are the set of states and actions. $\mathcal{R}(s, a)$ is the reward function when the agent is at state $s$ and takes action $a$. $\mathcal{S^{'}}$ is the set of possible next states reached from $\mathcal{S}$. As a baseline comparison to our method, we use Proximal Policy Optimization (PPO) \cite{DBLP:journals/corr/SchulmanWDRK17} an on-policy based method, which is the state of the art policy gradient method on continuous control tasks.

%cmaes mdp assumptions.. our method achieves a reward of 20
Unlike the MDP based methods, we use black-box based evolutionary strategies for policy search. CMA-ES, one of the most widely used variants of evolutionary strategy which is known for its robustness to multiple local minima and has proved to be efficient in solving RL problems in comparison to the state of the art policy gradient methods \cite{DBLP:journals/corr/SalimansHCS17}. The core idea is to sample policy parameters from a parameterized distribution like Gaussian and evaluate those parameters on the environment. The mean $\mu$ and the covariance matrix $\Sigma$ of the Gaussian distributions is then updated using the $M_{best}$ (highest rewards in our case) parameters, in the most standard case is 25, over every $g^{th}$ generation. Although, the covariance matrix calculation during every generation is expensive as it is $\mathcal{O}(N^2)$, and might not be feasible for parameters more than 10000, there are some practical tricks to mitigate these complexities \cite{ha2017visual}.

\begin{equation}
\mu = \frac{1}{M_{best}}\sum_{k=1}^{M_{best}} \Theta_k
\end{equation}
\begin{equation}
\Sigma_{ij} = \frac{1}{M_{best}}\sum_{k=1}^{M_{best}}(\Theta^i_k - \mu^i)(\Theta^j_k - \mu^j)^T
\end{equation}

%es.. rename carla to the original Since CMA-ES already has a closed method *not proefessional* of adapting the covariance matrix, to fuse the gradient information with the policy update would be challenging. 
In the above equation, $\Theta_k$ is the $k^{th}$ sampled parameter and $\Theta^i$ is the $i^{th}$ index of all sampled parameters. $\Sigma_{ij}$ refers to the $i^{th}$ column and $j^{th}$ row in a matrix. In the remainder of the paper, parameters refer to that of the neural network after flattening. We use the vanilla version of the ES algorithm, which involves updating the mean of the Gaussian distribution alone and having the covariance fixed. We improve this by using the dynamics to estimate the covariance matrix. Vanilla-ES method is based on finite differences and falls roughly into the category of \textit{Zeroth order} methods, which is used in applications, where gradients are not available. Compared to CMA-ES, these methods supposedly fail in a lot of applications, because of their inability to escape local minima. Similar to CMA-ES, these methods sample perturbations from standard Gaussian $\mathcal{N}(\mu, \sigma^2 I)$, where $I$ and $\sigma$ represents identity matrix and variance respectively, and use the evaluated values to move the mean towards a direction. We use antithetic sampling, which involves sampling $P$ points and $P$ mirror images of those points with respect to the mean of the distribution. Antithetic sampling has proven to reduce the variance of the \textit{Monte-Carlo estimator} as it draws correlated rather than independent samples \cite{DBLP:conf/icml/RenZE19}. All the members in a \textit{population} are then generated by $\mu + \epsilon \sigma$. These members are evaluated in the simulator. In the remainder of the paper, the term population represents a set of sampled policy parameters.

%notations.. weighting.. discussion of state vector
Now we present the notations and definitions which are used in this paper. We denote $\pi_{\theta}(s_t)$ as the policy which is a neural network, parametrized by $\theta$, which receives a state $s_t$ to output $a_t$ at every time-step $t$. To generate actions for each state using intermediate waypoints, apart from the policy, we also have the inverse dynamics model $\mathcal{M}_{\phi}(s_t, \hat{s}_t)$, which is parametrized by $\phi$. During the train-time, we also have access to the local waypoint $\hat{s}_t$, along with a state $s_t$. The gradient of the policy is represented by $\nabla \mathcal{L} (\theta)$, where $\mathcal{L}$ is the loss function, in our case is the Mean squared error of the actions predicted by the policy $\pi_{\theta}$ to that of the inverse dynamics model $\mathcal{M}_{\phi}$. $n$ is the total number of parameters in the policy neural network, and $N$ is the total number of workers in our distributed setup. Lastly, note that the bold counterpart of a symbol corresponds to an array of scalars or vectors, depending on that symbol.\\

%%IMPORTANT.. CHECK WHAT IS M AND H.. check blackbox
%%global wayopints and local waypoints.. trained model for expert action comparision accuracy
%Since all the trajectories are represented in Cartesian space, we can Euclidian distance metric to sample a local waypoint which is closer to the global waypoint and reachable by the agent as shown in the figure.
Also note that there are global waypoints denoted by $s^{'}_t$ which form the basis for the reward function, i.e., the agent gets a positive reward when it passes through a global-waypoint. The global waypoint is also used to estimate the heading angle in the state vector which the agent receives at every time-step. Since the global waypoint can have a varying resolution, i.e., sparser or denser, local waypoints are computed so that the agent can reach the next local waypoint and get closer to the global waypoint. During train-time, the agent has access to these local waypoints but not during test-time, which corroborates the use of inverse dynamics model only during training. Please note that $M$ is the max number of frames which the agent can see, whereas the episode length or horizon $H$ is set by the simulator. Since, it might not be possible for the simulator to find episodes which are of fixed size, the simulator is given a range of values to find an episode.

%%%%%%%%%%%%%%%%%%%%%%%%%%%%%%%%%%%%%%%%%%%%%%%%%%% because of lots of data generated in the early iteration, the optimal is reached quickly
\section{Proposed method}
In this section, we outline the main contributions of our method. The detailed implementation details are given in the experiments section and note that our method along with the other evolutionary baselines are run on a cluster of machines, as a distributed, centralized setup, where a Master node generates the parameters and worker nodes evaluates them on the simulator.

%%%%%%%% define what is on-policy.. tensorflow for dynamics.. weighting for carla
\subsection{Estimating actions}
We use an inverse dynamics model $\mathcal{M}_{\phi}$ to estimate the action using the current state and the locally generated waypoint. As mentioned before, a simulator provides a global waypoint and a local waypoint is generated with the constraint of agent's reachability. At the end of each generation, with the exception of the first generation, we aggregate all the rollouts until the current generation and train the inverse dynamics model for every iteration. Since this process happens at different worker nodes in a cluster, each node only has access to its own history of rollouts, and this procedure occurs in parallel at different nodes.

Since the actions which are estimated by the inverse dynamics model corresponding to each state in a trajectory are on-policy, we can use behaviour cloning using DAgger \cite{DBLP:journals/jmlr/RossGB11} to fit the policy on the data collected. To achieve that, the worker estimates the gradient of the policy parameters by using the following Mean squared error (MSE) loss function and does gradient descent.

\begin{equation}
 \mathcal{L}(\theta) = \frac{1}{M}\sum_{j=0}^{j=M}(\pi_{\theta}(s_t) - \mathcal{M}_{\phi}(s_t, \hat{s}_t))^2
\end{equation}

In the above equation, $\pi_{\theta}$ is the policy, $\mathcal{M}_{\phi}$ is the inverse dynamics model, and $M$ is the episode length. After every generation, each worker would fit the policy $\pi_{\theta_i}$ using the aggregated data by taking successive gradient steps, after which we obtain $\theta^*_i$, where $i$ is the index of the worker. $\theta_i^*$ along with the cumulative reward $R_i$ are then sent to the master. The reader is reminded that, for the sake of simplicity, Unlike $\mathbf{\Theta_g}$ (array of policy parameters), $g$ (generation) is omitted out from $\theta_i$ (policy parameters received by $i^{th}$ worker. This is the same in case of $\mathbf{R_g}$ and $R_i$.

The parameters of $\mathcal{M}_\phi$ are updated at the master node by training on the trajectories recorded from all workers, during a particular generation. Another thread, meanwhile, waits for the rewards $\mathbf{R_g}$ and $\mathbf{\Theta_g}$, which are a vector of rewards $R_i$ and $\theta^*_i$ respectively. Both of them happen concurrently, to speed-up training.

%what is a population, rename all f to policy.. how many iteration do you train your policy.. tf for inv_dyn.. sgd or something.. value of beta
%%%%%%%% check antiethic spelling beta value is 1.0.. what is a generation.. subspace dimension kxk
\subsection{Incorporating gradients}
As stated in the preliminaries, for vanilla-ES algorithm, we sample some perturbations $\boldsymbol{\epsilon}$ around the mean of the Gaussian $\mu$, and update it as follows:

\begin{equation}
    \mu_{g+1} \leftarrow \mu_{g} - \gamma \cdot \frac{\beta}{\sigma^{2}2P} \sum_{i=0}^{P} \epsilon_i \cdot [F(\mu_g+\epsilon_i) - F(\mu_g-\epsilon_i)]
\end{equation}

In the above equation, $2P$ is the number of members in a population, since we are doing antithetic sampling. $F$ is the fitness function or the reward function. $\beta$ and $\sigma$ are the normalizing coefficient and variance, respectively. $\gamma$ is the learning rate of the gradient ascent (since we maximize the reward function). $\mathbf{\Theta}_g$ is an array of policy parameters, that are sent to the workers are estimated using $(\mu + \boldsymbol{\epsilon})$ ($\mu$ would get broadcasted over the array $\boldsymbol{\epsilon}$). $\epsilon_i$ is a perturbation sampled from $\mathcal{N}(0, I_n)$, where $n$ is the length of the policy parameters and so the identity matrix $I_n$ is of dimension $n\times n$.

Recently, \cite{DBLP:conf/icml/Maheswaranathan19} proposed a method to improve training of the Vanilla-ES in situations where surrogate/noisy gradient is available. Instead of having a search covaraince of $\sigma I_n$, the solutions could then be drawn from a Gaussian having $\Sigma$ as follows:

\begin{equation}
\label{eq:sample}
 \Sigma = \frac{\alpha}{n} I_n + \frac{1-\alpha}{k} U U^T
\end{equation}

%substracting the updates from the updated mean... hyperlink all the figures to the text.. surrogate is not mentioned properly in the text
In the above equation, $U$ is a low-rank, orthogonal basis of the subspace spanned by the surrogate gradients. The orthogonal basis can be obtained easily from QR decomposition of the gradients, which is an $k\times n$ matrix, where $n$ is the number of policy parameters/ size of the gradient and $k$ is the number of gradients obtained, which is equal to the number of workers. $\alpha$ is the coefficient which weights the gradient perturbations over the random ones. Taking inspiration from \cite{DBLP:conf/icml/Maheswaranathan19}, we also apply a similar concept in our case, where the surrogate gradients could be pointing towards the optimal parameters. From the previous subsection, we could use:

\begin{equation}
    \label{eq:surr}
    \mathbf{\Delta_g} = \frac{\mathbf{\Theta_g}^* - \mu_{g+1}}{\eta}
\end{equation}

%intution why this might be helpful... multiple robots.. i index is always used for workers.. M is the number of max frames.. N are the number of workers
In the above equation, $\mathbf{\Delta_g}$ represents an array of surrogate gradients. Again, since $\mu_{g+1}$ is a single vector and $\mathbf{\Theta_g}^*$ is an array of vectors ($\theta^*_i$), by broadcasting we get an array of surrogate gradients. The normalization constant $\eta$ is to ensure that the covariance matrix does not become large, to make the sampling inaccurate. The CMA-ES algorithm does not explicitly control the step-size of the distribution, but at any given step or generation, it increases or decreases the scale only in a single direction for each selected step \cite{DBLP:journals/corr/Hansen16a}. But in our case, since the gradient subspace consists of multiple optimum points, we can induce the distribution towards the direction of the optimal policy, as shown in Figure \ref{fig:specific}. The entire algorithm is described in the figure on this page, with steps corresponding to master or worker nodes in blue and red, respectively.

%%%%%%%%%%%%%%%%%%%%%%%%%%%%%%%%%%%%%%%%%%%%%%%%%%%%%%%%%%%%%%%%%%%%%%%%%%%%%%%%%%%%%%%%%%%%%%% dimension of I underscript k
\begin{algorithm}
\caption{Policy shaping using surrogate gradients}
\begin{algorithmic}[1]
\REQUIRE Master and worker steps in blue and red
\REQUIRE $U$ as the gradient subspace
\REQUIRE $\Delta$ as the surrogate gradient
\REQUIRE $N$ as the number of workers
\REQUIRE $\{\mathcal{D}_g^i\}_{i=0}^{N}$ as data-buffers for each worker
\STATE Initialize $\{\mathcal{D}_i\}_{i=0}^{N}$ to empty set
\STATE Initialize $\mu_0$ to $0$
\STATE Initialize $U$ to $n \times n$ identity matrix

\WHILE{not done}
\color{blue}
%\STATE Sample $\boldsymbol{\epsilon_n} \sim \mathcal{N}(0, \sigma I_n)$ and $\boldsymbol{\epsilon_k} \sim \mathcal{N}(0, \sigma I_k)$
\STATE Estimate $U$ from {\tt\small QRDecompose(}$\mathbf{\Delta_g}${\tt\small)}
\STATE Estimate solutions $\mathbf{\Theta_g} \sim \mathcal{N}(\mu_g, UU^{T})$
\STATE {\tt\small comm.send($\mathbf{\Theta_g}$)}

\color{red}
\STATE {\tt\small comm.recv($\theta_i$)}
\STATE Evaluate $\theta_i$ to obtain $\{{s_j, \hat{s}_j}\}_{j=0}^{M}$
\STATE Save $\{s_j, \hat{s}_j, a_j\}_{j=0}^{M}$ onto NFS
\STATE Aggregate data: $\mathcal{D}_g^i \leftarrow \{s_j, \hat{a}\}_{j=0}^{M} \cup \mathcal{D}_{g-1}^i$
\STATE Estimate $\hat{a}$ from $\mathcal{M}_{\phi}(s_j, \hat{s}_j)$
\STATE Train on $\mathcal{D}_g^i$ to obtain $\theta_i^*$
\STATE {\tt\small comm.send(${R_i}$, $\theta_i^*$)}

\color{blue}
\STATE {\tt\small comm.recv($\mathbf{R_g}$, $\mathbf{\Theta_g^*}$)}
\STATE $\mu_{g+1} \leftarrow \mu_g - \gamma \cdot \frac{\beta}{2\sigma^{2}P} \sum_{i=0}^{P} \epsilon_i \cdot [F(\mu_g+\epsilon_i) - F(\mu_g-\epsilon_i)]$
\STATE Calculate $\mathbf{\Delta_g} = \frac{\mathbf{\Theta_g}^* - \mu_{g+1}}{\eta}$
\STATE Train $\mathcal{M}_\phi$ using experiences collected

\ENDWHILE
\end{algorithmic}
\end{algorithm}

%=============================================================================== Further, since our method involves generating actions for a pair of agent's state and waypoint
%(\url{https://www.open-mpi.org/})
%(\url{https://hub.docker.com/})
\section{Experimental Results}
Our method was evaluated on two distinct simulators involves long horizon tasks with sparse rewards. At this point, we would like to remind the reader that our method is not only limited to simulators; we could use the same setup in cases where there are multiple robots to speed-up training. Many popular works use simulators like OpenAI \cite{DBLP:journals/corr/BrockmanCPSSTZ16} and Mujoco \cite{DBLP:conf/iros/TodorovET12}, where the simulations are run as a wrapper inside the worker scripts which could be parallelized. Unlike them, our entire experimental setup was based on a distributed client-server setup. We use docker containers to run the simulations, as our simulators involve high-end graphics and so have to be run as standalone servers. We used 13 machines of different hardware configurations and set them up as a Beowulf cluster. We use a common Network File system (NFS) shared amongst all the machines in the cluster, onto which all the trajectories are written to, and are used by the master node to train the inverse dynamics model. All the distributed methods were implemented in OpenMPI. Experimental details irrespective of the simulator used are as follows.

%common training details
At the end of every generation, the fitness/reward values are rank transformed to the range of -1 to +1 to eliminate any outliers in the population dominating the update procedure. In the case of CMA-ES, weight decay of 0.05 was used to prevent overfitting. We used a population size of 64, evaluated by 64 train workers. Additionally, we used 15 evaluation workers to evaluate the best member of the population. Each worker evaluates the weight vector two times using different seeds to obtain the mean cumulative reward and the gradients. For both the simulators, we followed a similar reward strategy similar to the CarRacing-v0 simulator from the OpenAI gym \cite{DBLP:journals/corr/BrockmanCPSSTZ16}. The agent receives a reward of $(\frac{M}{H})$ if it passes through a waypoint, where $M$ and $H$ are the max-frames and horizon, respectively. In case of a collision, a negative reward of -50 is received and the episode is terminated. To encourage the agent to reach the goal faster, it also receives a negative reward of -0.1 irrespective of any state (positive or negative rewarded state). In regards to the training of the inverse-dynamics model, trajectories which do not have a cumulative positive reward of 5 are discarded.

%$\alpha$ in equation \eqref{eq:sample} and $\eta$ in equation \eqref{eq:surr} was set to 0.6 and 0.01 respectively.
%in the case of navigation are more accurate as the dimension is smaller and the dynamics are easier compared to the robot arm environment.

\begin{figure}
    \includegraphics[width=\columnwidth]{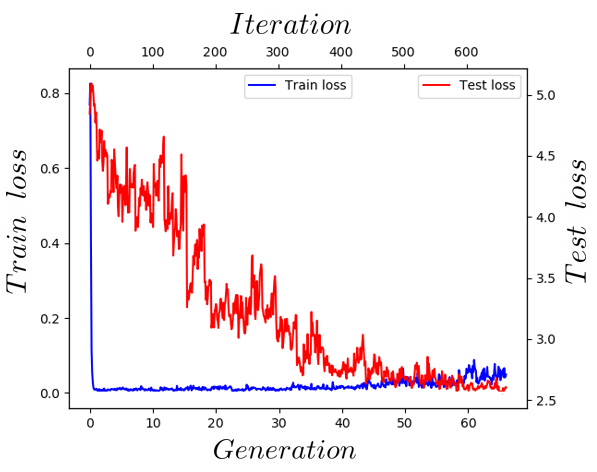}
    \caption{Accuracy of the inverse dynamics model ( $\mathcal{M}_\phi$) with generation in the Carla simulator. $x$ axis represents the training iteration which is folded in with the generation. Test Loss corresponds to the euclidean distance between the actions and the ground-truth actions. Overall, we can see that the accuracy of the action prediction improves with iteration. Notice how the train loss increases slowly. This is because the data collected to train $\mathcal{M}_\phi$ gets aggregated after each generation and so the loss increases because there are more samples to fit.}
    \label{fig:dyn}
\end{figure}

\subsection{Carla Driving simulator}
We used the Carla simulator \cite{DBLP:conf/corl/DosovitskiyRCLK17}, which is a High-end graphical simulator based on Unreal Engine. The task was to navigate from a source to destination location which are both sampled randomly in a specific environment/town (Town07). The agent receives a 38-dimensional state consisting of heading angle, odometry information, linear and angular velocity of the vehicle and the 32-dimensional lidar scan at every time-step. The agent then outputs an action vector of 2 dimensions, i.e., steering and throttle. To improve stability, we set a speed limit of 25 Kmph (whenever the agent crosses it, a brake is automatically applied). The resolution of the rewards is also a hyperparameter, which is set 1.5. As the resolution increases, the rewards tend to become more sparse. Performance of our method in comparison with other benchmarks are shown in Figure \ref{fig:benchmark}.

%inv_dyn for carla.. check the m and h ... Consequently***?, the agent would not get positive rewards exactly at every step, even though it might execute the trajectory optimally
All the episodes had the number of waypoints between 800 and 1200. The maximum number of frames which the agent could see in a specific episode was set to 1000. Based on these values, when the agent is within the radius of 0.5m of the next waypoint, it receives a reward of 3.16. Other than this, there are no other positive rewards, which the agent could receive. The inverse dynamics model $\mathcal{M}_\phi$ is a two-layer neural network with 20 neurons in each layer with ReLU activation at each layer. We used a learning rate of 0.001 for training and a batch size of 64. Train-loss and test-loss of the inverse dynamics model $\mathcal{M}_\phi$ of the Carla simulator are shown in Figure \ref{fig:dyn}. We use the actions from a fully trained policy as the ground-truth data. Train-loss corresponds to the loss obtained when training $\mathcal{M}_\phi$ on the rollouts. We compute Euclidean distance between the actions generated by $\mathcal{M}_\phi$ and the ground-truth actions, as the test-loss.

%\begin{equation}
% \mathbf{B(t)} = \sum_{i=0}^{n}\binom{N}{k} (1-t)^{n-i}t^i\mathbf{P}_i,\hspace{0.5cm} 0 \leq t \leq 1
%\end{equation}

\subsection{UR5 Robot arm in ROS Gazebo}
Similar to the Navigation task mentioned above, we further tested our method on a robotic arm simulator developed by \cite{DBLP:journals/corr/abs-2007-02753}. The task was to make the end-effector reach a goal coordinate from a fixed configuration (C-space), as shown in Figure \ref{fig:intro}. We modified the environment by computing Bezier curves from the source to the target end-effector position, which is randomly sampled from the upper hemisphere. We used a quadratic Bezier curve which uses a second-order Bernstein polynomial involving one control point $\mathbf{P_1}$ which in our case was $(.75, 0.15, 0.45)$, apart from the source ($\mathbf{P_0}$) and the target ($\mathbf{P_2}$) end-effector poses. All the points belonging to this curve are obtained using different values of $t$ ranging from 0 to 1. The reward mechanism was similar to the one used in the navigation task, i.e., -0.1 for every time-step and positive reward if the end-effector is within a distance of 0.1 units from the next local waypoint. In total, our method was able to achieve a cumulative reward of -2.8 after training for 200 generations, as shown in Figure \ref{fig:benchmark}. Unlike the Carla simulator, since we now have access to the exact number of points which we can sample from a Bezier curve, the horizon was equal to the maximum number of frames which was set to 100. A similar inverse dynamics model from the Carla simulator was used here, with the modification of adding another layer, while keeping all the other details same. The input to the inverse dynamics model are joint velocities, Cartesian coordinates of the next waypoint with respect to the end-effector position in 3D space. The action vector consists of 5D joint torques.

%Accuracy tells whether the predicted action value is in the same direction as of the ground-truth actions.

%For example, in the case of steer command ..
%measuring the accuracy... change the inverse dynamics model to mathcal M
%We also experimented by sparsifying the rewards in the environment. As mentioned before, .. We then make the waypoints even more sparse, keeping all the other values fixed. We do this by decreasing the relative resolution of the global planner in CARLA by one-tenth. In which case, the agent would obtain a reward of 36.2 when it get's near each waypoint.

\section{Discussion and Conclusion}
%what is this paper about
In this paper, we presented a method which could improve the performance of distributed evolutionary strategies for RL in scenarios where we have access to waypoints. Our method is tested on the Carla driving simulator and the UR5 Gazebo simulator and the results indicate the validity of the hypothesis. By incorporating generated actions using an inverse-dynamics model, we have shown that the training process can be improved without additional cost, as the experiences are generated during training. This method could also be used in real-world scenarios when we have multiple robots. In other cases, it could also be used in \textit{Sim2Real} transfer scenarios, by training the policy in a simulator, where we have access to data in abundance.

In many real-world scinarios, we cannot obtain waypoints directly, unlike simulator. However, there are many global planners which allow us to do that. For example in visual navigation, we could run A* search at every time step to find out the next closest waypoint to reach the goal destination. In such a case, it must be ensured that these waypoints generated by a standard planner should be within the same distribution as the state transitions, to avoid any covariate shifts for the inverse dynamics model.

%what kind of experimental setup

%\begin{figure}
%    \centering
%    \includegraphics[width=\columnwidth]{placeholder.jpeg}
%    \caption{Performance of different methods with sparser rewards on CARLA driving simulator. When we change the resolution of the global waypoints to make them more sparse, while keeping the maximum obtainable reward same, the performance of the baselines worsens. Since our method enables the policy learn the local behaviours, from the gradients evaluated using the generated actions, it performs better in these situations.}
%    \label{fig:sparse}
%\end{figure}

%\section*{ACKNOWLEDGMENT}

%This work was supported by C-BRIC
%(one of six centers in JUMP, a Semiconductor Research
%Corporation (SRC) program sponsored by DARPA), National Science Foundation (grants CCF-1317433 and CNS-1545089) and Intel Corporation. The authors affirm that the views expressed herein are solely their own, and do not represent the views of the United States government or any agency thereof.

%\bibliographystyle{unsrt}
\printbibliography

\end{document}